\documentclass[11pt, conference]{IEEEtran}
\pdfoutput=1
\usepackage{orcidlink}
\usepackage{silence}

\usepackage{amsmath,amssymb,amsfonts}
\usepackage{graphicx}
\usepackage{pifont}

\usepackage{amsthm}
\usepackage{xcolor} 
\usepackage[figuresright]{rotating}

\usepackage{graphicx}

\usepackage{array}
\usepackage{textcomp}
\usepackage{xcolor}
\usepackage{multirow}
\usepackage{booktabs}
\usepackage{enumitem}

\usepackage{mathtools}
\usepackage{breqn}
\usepackage{float}

\usepackage{pgfplots}
\pgfplotsset{compat=1.18}
\usepgfplotslibrary{groupplots}

\usepackage[font=scriptsize]{caption}
\usepackage{subcaption}

\usepackage{multirow,tabularx}
\usepackage{hyperref}
\usepackage{flushend}
\usepackage{algorithmic}

\usepackage[vlined, ruled, shortend,,linesnumbered]{algorithm2e}

\SetCommentSty{mycommfont}

\usepackage{pgfmath}
\usetikzlibrary{decorations.text, arrows.meta,calc,shadows.blur,shadings}

\newlength\figureheight
\newlength\figurewidth
\SetAlCapNameFnt{\footnotesize}
\SetAlCapFnt{\footnotesize}

\title{
    Fusing Odometry, UWB Ranging, and Spatial Detections for Relative Multi-Robot Localization
}

\author{
    \IEEEauthorblockN{
        \vspace{1em}
        Xianjia Yu\IEEEauthorrefmark{2}\,\orcidlink{0000-0002-9042-3730},
        Iacopo Catalano\IEEEauthorrefmark{2}\,\orcidlink{0000-0001-9212-8615},
        Paola Torrico Mor\'on\IEEEauthorrefmark{2}\,\orcidlink{0000-0002-1253-4574},\\ [-12pt]
        Sahar Salimpour\IEEEauthorrefmark{2}\,\orcidlink{0000-0003-4954-0000},
        Tomi Westerlund\IEEEauthorrefmark{2}\,\orcidlink{0000-0002-1793-2694},
        Jorge Pe\~na Queralta\IEEEauthorrefmark{2}\IEEEauthorrefmark{3}\,\orcidlink{0000-0003-3091-3217}
        \vspace{1em}
    }
    \IEEEauthorblockA{
        \normalsize
        \IEEEauthorrefmark{2}\href{https://tiers.utu.fi}{Turku Intelligent Embedded and Robotic Systems (TIERS) Lab, University of Turku, Finland}. \\
        \IEEEauthorrefmark{3}\href{https://scai.ethz.ch/}{SCAI Laboratory at SPZ, Swiss Federal School of Technology in Zurich - ETH Zurich, Switzerland}.\\
        Emails: \textsuperscript{1}\{xianjia.yu,imcata, pctomo, sahars, jopequ, tovewe\}@utu.fi, jorge.penaqueralta@hest.ethz.ch\\[+6pt]
    }
}

\begin{document}

\maketitle
\thispagestyle{empty}
\pagestyle{empty}



\begin{abstract}\label{sec:abstract}%
    This letter presents a cooperative relative multi-robot localization design and experimental study. We propose a flexible Monte Carlo approach leveraging a particle filter to estimate relative states. The estimation can be based on inter-robot Ultra-Wideband (UWB) ranging and onboard odometry alone or dynamically integrated with cooperative spatial object detections from stereo cameras mounted on each robot.
    %
    The main contributions of this work are as follows.
    First, we show that a single UWB range is enough to estimate the accurate relative states of two robots when fusing odometry measurements. Second, our experiments also demonstrate that our approach surpasses traditional methods, namely, multilateration, in terms of accuracy. 
    Third, to further increase accuracy, we allow for the integration of cooperative spatial detections. Finally, we show how ROS\,2 and Zenoh can be integrated to build a scalable wireless communication solution for multi-robot systems. The experimental validation includes real-time deployment and autonomous navigation based on the relative positioning method.
    It is worth mentioning that we also address the challenges for UWB-ranging error mitigation for mobile transceivers. 
    conditions. 
    The code is available at \href{https://github.com/TIERS/uwb-cooperative-mrs-localization}{https://github.com/TIERS/uwb-cooperative-mrs-localization}.

\end{abstract}

\IEEEpeerreviewmaketitle


\section{Introduction}\label{sec:introduction}
Ultra-wideband (UWB) wireless radio communication and ranging technology has been increasingly infiltrating the domain of mobile robotics, as well as finding wider applications in the Internet of Things(IoT), Augmented Reality, and industrial use cases among others~\cite{kim2022uwb}.
Since UWB ranging sensors offer low-cost and centimeter-level out-of-the-box accuracy, they have gradually gained attention in autonomous systems applications, including UWB-based state estimation with and without fixed infrastructure, and UWB mesh sensor networks~\cite{xianjia2021applications}. This technology holds significant potential for relative state estimation in multi-robot systems~\cite{xu2022omni, xun2023crepes}, a crucial yet still challenging research topic in GNSS-denied environments and outdoors, where GNSS signals may degrade~\cite{xianjia2021cooperative, zhang2019combined}. Moreover, it can serve as the basis for collaborative tasks 
such as search and rescue, and terrain inspection~\cite{xu2020decentralized, queralta2020collaborative}. Additionally, this approach can be extended to transitions of multi-robot systems between indoor and outdoor scenarios.

\begin{figure}[t]
    \centering
    \includegraphics[width=0.45\textwidth]{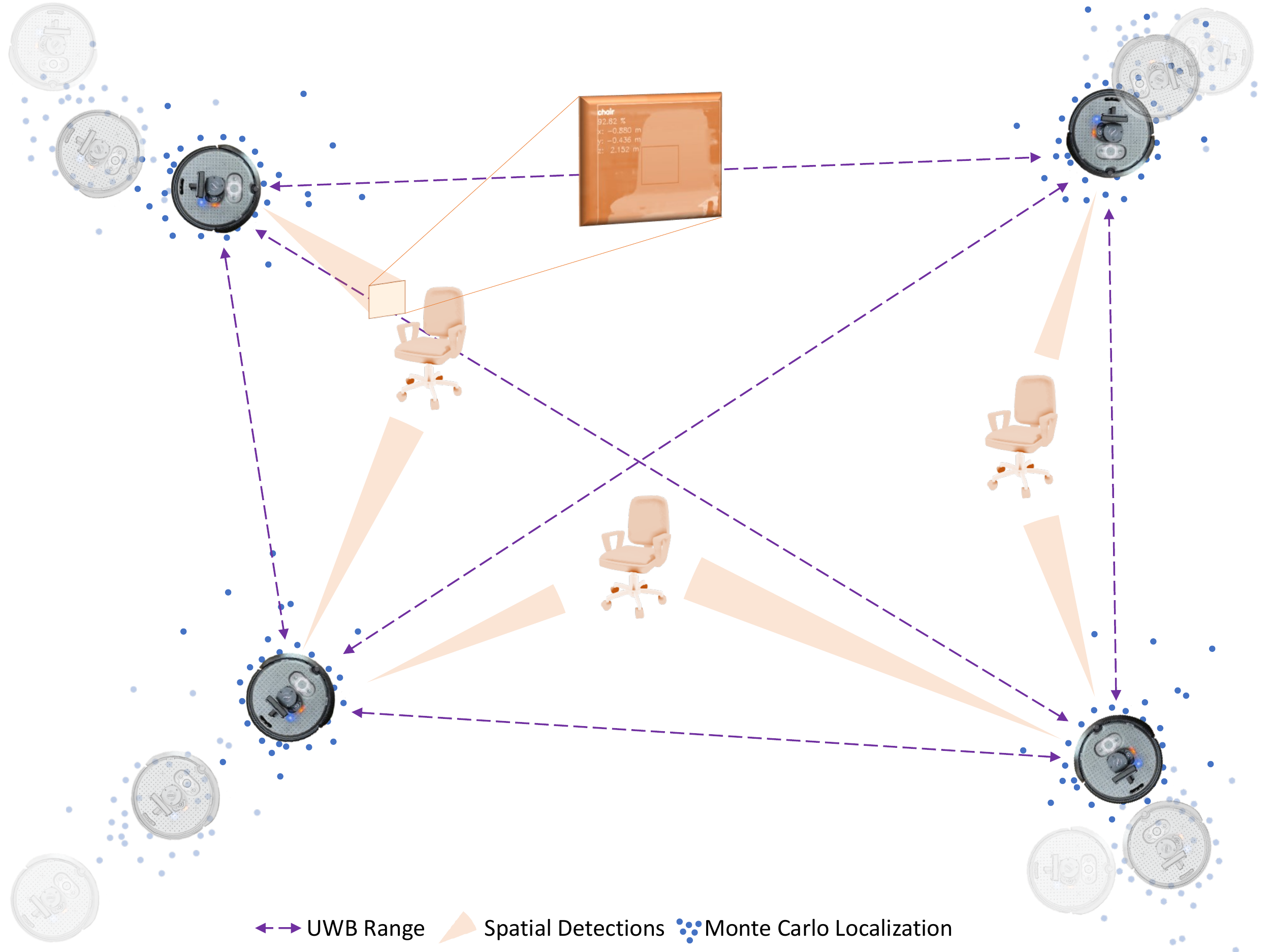}
    \caption{Conceptual diagram of the proposed particle filter-based multi-robot relative localization fusing UWB ranges, robot odometry, and cooperative spatial detections.}
    \label{fig:pf_diagram}
\end{figure}

State estimation based on the fusion of UWB and other sensors or estimators, such as LiDAR odometry (LO)~\cite{shan2018lego} and visual-inertial odometry (VIO)~\cite{he2020review}, has been the subject of numerous research efforts~\cite{xu2022omni, xu2020decentralized, nguyen2021viral}. LO and VIO, which focus on ego-state estimation, are among the most popular and reliable state estimation techniques. However, porting these approaches to multi-robot systems is often either computationally expensive or complex to achieve in realistic scenarios~\cite{cramariuc2022maplab, tranzatto2022cerberus}. Either independently or as part of collaborative SLAM processes, multi-robot localization methods often use environment features to adjust for potential drift or error in the relative state estimation~\cite{cao2021vir}, or the optimization of map matching or pose graphs~\cite{boroson2020inter}.



In our approach, we aim to provide a flexible and cost-effective relative multi-robot localization method. Instead of using additional sensors for robots sensing each other's position or optimizing raw sensor data matching, we leverage simultaneous visual sensing of objects in the environment using stereo cameras – a common sensor in mobile robots. We refer to this method as cooperative spatial detections, where objects in the environment are simultaneously detected by multiple cameras, and their 3D pose is identified with respect to each robot's sensor.

Our motivation is to provide a flexible relative multi-robot localization method that can fuse odometry and ranging measurements, with a minimum of just a single range, but also incorporate cooperative spatial detections when available. We, therefore, propose
%
a sequential Monte Carlo optimization approach to calculate the relative position among robots with a particle filter (PF). Fig.~\ref{fig:pf_diagram} shows a conceptual illustration of the working mechanism and sensing modalities in the proposed approach. Initially, We develop this approach and explore its potential with a multi-robot system of ground robots. We also validate the performance of the proposed approach comparing it to the conventional UWB relative positioning approach based on multilateration. Owing to various sources of UWB ranging error, multiple studies have shown the importance of mitigating the UWB ranging error as part of the workflow of UWB data processing~\cite{zhao2021learning}. To achieve this, we use a Long Short-Term Memory (LSTM) network to calibrate the UWB range data 
prior to the PF deployment. In contrast to previous studies, we evaluate multiple types of LSTM networks, considering both UWB measurements and robot orientation.

Another key element concerning existing works in infrastructure-free UWB-based relative positioning is that we assume that only a single UWB transceiver is available for each robot. However, the same methods can be applied with multiple transceivers per robot. This is, to the best of our knowledge, the first infrastructure-free UWB localization system that is generalizable to an arbitrary number of nodes and robots, and that integrates odometry and cooperative detections.
We adopt a shared two-dimensional yaw-orientation reference for all robots, where the orientation is not computed relatively between robot pairs but rather with respect to a unified global reference frame. Consequently, at the beginning of the experiment, all robots are initialized with identical orientations.
In summary, the main contributions of this work are the following:
\begin{enumerate}[label=\roman*)]
    \item a novel sequential Monte-Carlo relative localization method fusing odometry and ranging measurements, requiring a minimum of just one range;  
    \item the addition of an LSTM network for reducing individual range errors calibrated for individual pairs of UWB transceivers (all moving) by taking not only the UWB measurements but also the orientation of the robot as input;
    \item the integration of cooperative spatial detections to further increase the accuracy of the relative state estimation;
    \item the demonstration of a practical multi-robot deployment with ROS\,2 and Zenoh, along with the consistency of the particle filter at higher dimensions, exhibiting computational requirements similar to a baseline multilateration implementation.
\end{enumerate}

The rest of this paper is organized as follows. Section~\ref{sec:related_work} mainly discusses related work about UWB relative state estimation systems in a multi-robot system. Section~\ref{sec:methodology} then introduces the proposed approach and the experimental settings for verification.  Experimental results are shown in Section~\ref{sec:experiment}. Finally, Section~\ref{sec:conclusion} concludes the paper.



\section{Related Work} \label{sec:related_work}
This section covers the state-of-the-art UWB-based relative localization and high-accuracy methods for UWB ranging. 

\subsection{UWB for cooperative positioning}

The majority of UWB positioning systems are based on ranging between a mobile node, or tag, and a set of fixed nodes in known locations, or anchors. This is the case of commercial, out-of-the-box systems as well as most UWB-based localization methods in the literature~\cite{queralta2020uwb}. We are however more interested in infrastructure-free relative localization methods where all UWB transceivers are potentially mounted on mobile robots. This approach has the benefit of being more flexible from an ad-hoc deployment perspective. In multi-robot systems, infrastructure-free localization can also 
significantly facilitate the positioning transition for robots from indoor to outdoor scenarios. The authors in~\cite{guler2020peer} adopted a Monte Carlo localization approach to compute the relative localization between two aerial robots. This involved attaching one UWB tag to a mobile robot and placing three UWB anchors on another robot in a stationary position. We propose here a more general and scalable approach, which naturally adapts to a variable number of UWB nodes.

More conventional approaches include multilateration with least squares estimators~\cite{xianjia2021cooperative} and different extended Kalman filter (EKF) approaches depending on the sensor data being fused~\cite{nguyen2021viral}. Alternatively, machine learning (ML) approaches, including LSTM networks, have also been proposed~\cite{poulose2020uwb}. Some of the state-of-the-art works fuse UWB with other sensors and estimators (IMU, LO, or VIO) through sliding window optimization methods~\cite{xu2022omni, xu2020decentralized}. However, to the best of our knowledge, current methods achieving high accuracy still require other estimators (e.g., lidar or visual odometry) or need a higher number of ranges to be measured (either for each pair of robots or with a higher number of robots).

In this work, we use instead a particle filter as the problem of multi-robot localization with a single UWB node in each robot results in a non-Gaussian optimization problem with multiple local minima. This is due to the multiple spatial realizations for different subsets of the localization graph, with methods such as an EKF potentially converging to local minima without convergence guarantees to a globally optimal solution.




\subsection{UWB ranging error mitigation}

Despite the superior performance of UWB ranges in comparison with other wireless radio technologies, there are still different sources of measurement errors.
The most common include multi-path propagation interference, electrical interference, or thermal noise. 
It is challenging but essential to mitigate the error in individual ranging measurements before performing state estimation using these measurements as the key component. 
Several studies adopt ML or deep neural network methods to model the error~\cite{wang2020semi, ridolfi2021uwb, fontaine2020edge}. 
In~\cite{ridolfi2021uwb}, an algorithm that combines ML and the time resolution capabilities of UWB with adaptive physical settings was proposed to enable the automatic calibration of the anchor positions and decrease the ranging error.
Alternatively, Wang et al.~\cite{wang2020semi}a semi-supervised support vector machine (SVM) was proposed to identify the NLOS and mitigate the ranging error by $10\%$. Additionally, in~\cite{fontaine2020edge}, researchers demonstrated that autoencoders could achieve a decent performance of estimation accuracy.

\subsection{LSTM networks in UWB positioning system}

LSTM networks are especially suited to time-series data, with UWB ranging measurements being a clear example. Previous works on the application of LSTM have primarily focused on anchor-based UWB positioning systems. In one case~\cite{poulose2020uwb}, researchers have applied LSTM  to directly estimate the user position with anchors, which achieved competitive accuracy in a simulated environment with respect to the conventional trilateration.
In~\cite{wang2020uwb}, the authors applied an LSTM to estimate the fixed anchor based UWB ranges based on time of arrival (TDoA), relying solely on prior UWB measurements. 
By classifying channel conditions with channel impulse response of the received UWB signal with LSTM, in~\cite{kim2022uwb}, the authors mitigated the positioning degradation caused by the NLOS situations and obtained accuracy improvement for the localization based on the EKF.

\subsection{\texorpdfstring{ROS\,2}2 and Zenoh}
ROS\,2 is seeing increasing adoption, driven by inherently better distribution, abstraction, asynchrony, and modularity~\cite{macenski2022robot}. Although the clear contribution of these factors to decentralized multi-robot systems, multiple challenges remain in scaling up communication~\cite{zhang2022distributed}. The communication middleware for ROS\,2 is highly dependent on the discovery system provided by the Data Distribution Service (DDS) protocol. 
However, the out-of-the-box performance of DDS over wireless networks is unreliable with a large number of ROS nodes and devices when peer autodiscovery is enabled. 
A promising and more scalable solution is Zenoh, developed under the Eclipse foundation. 
Zenoh is a protocol that claims to have reduced by up to \(99.97\,\%\) the discovery traffic compared with other DDS communications. In our experiments, we have seen significant qualitative performance and reliability improvements,  as well as reduced latency, and increased bandwidth. However, it is out of the scope of this paper to delve into a quantitative analysis of the improvements Zenoh brings over DDS.


\section{Methodology}\label{sec:methodology}

\begin{figure}
    \centering
    \includegraphics[width=0.49\textwidth]{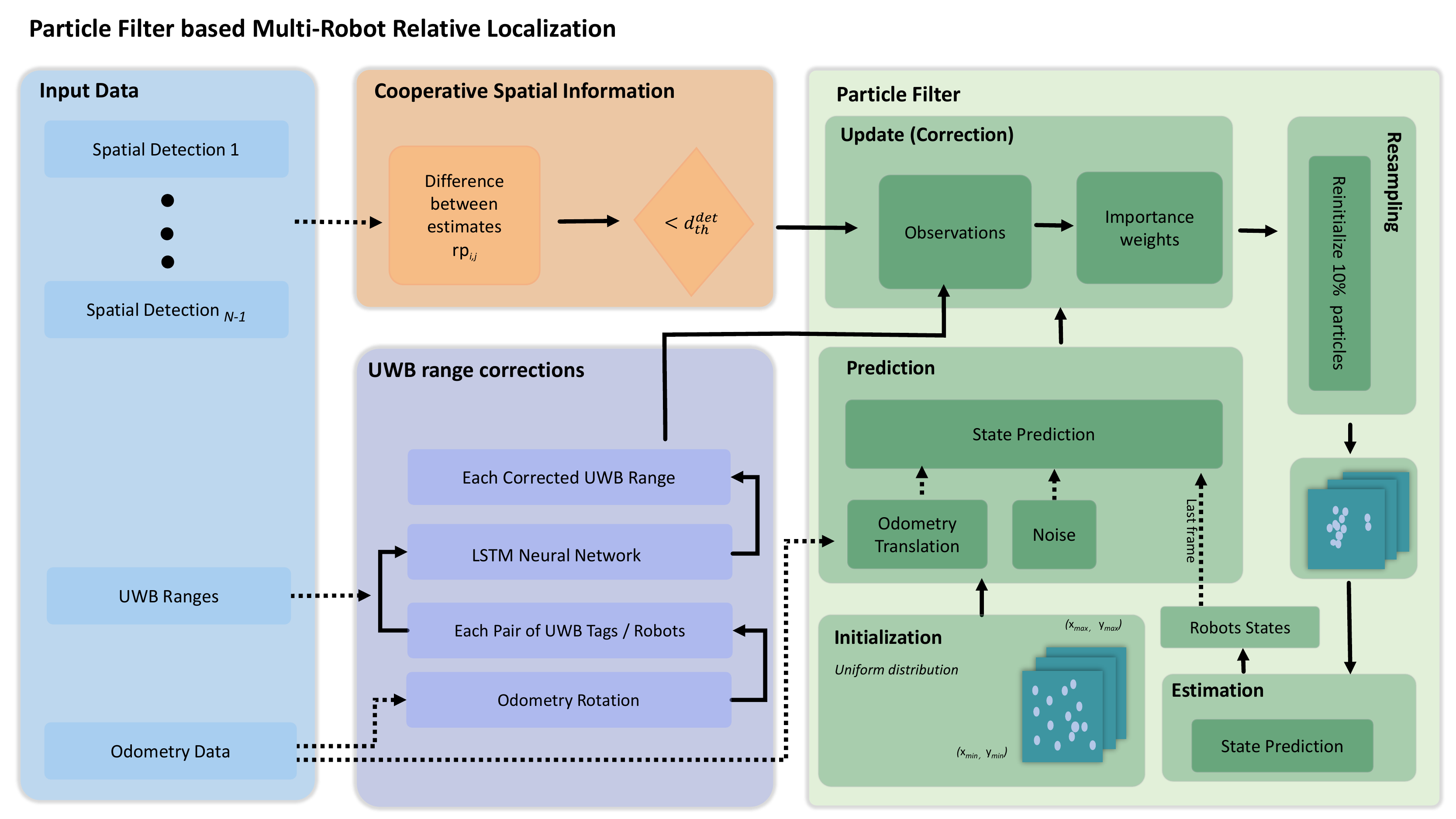}
    \caption{The detailed process (from left to right) of the particle filter-based multi-robot relative position system where \(d^{det}_{th}\) is the threshold determining if detections from different robots correspond to the same object.}
    \label{fig:uwb-ral-diagram}
\end{figure}

In this section, we introduce the relative localization problem, which is formulated as an optimization task using inter-robot UWB ranging measurements, individual odometry measurements, and cooperative spatial detections as inputs. These inputs are conceptually depicted in Fig.~\ref{fig:pf_diagram} and the workflow is demonstrated in Fig.~\ref{fig:uwb-ral-diagram}. 

\subsection{Problem Statement} 

We use the following notation for the remainder of this paper. We consider a group of $N$ robots, or agents, with positions denoted by $\textbf{p}_i(t)\in\mathbb{R}^3$ in a common reference frame, with $i\in\{1,\dots,N\}$. Agents can measure their relative distance to a subset of the other agents in a bidirectional way, i.e., with both agents calculating a common ranging estimation simultaneously. The set of robots and the estimated distances between them are modeled by a graph $\mathcal{G}=(\mathcal{V},\mathcal{E})$, where $\{1,\dots,N\}$ is a set of $N$ vertices, and $\mathcal{E} \subset \mathcal{V}\times \mathcal{V}$ is a set of $\lvert E \rvert \leq N(N-1)/2$ edges. We consider undirected graphs, i.e., $(i,j) \in \mathcal{E} \Longleftrightarrow (j,i) \in \mathcal{E}$. It is worth noting that the graph does not need to be fully connected, and only a subset of range measurements is needed.

We consider robots in three-dimensional space. Nonetheless, for the purpose of collaborative localization, and without a loss of generality, we assume that the position of each agent is given by $\textbf{p}_i\in\mathbb{R}^2$. Even if a locally rigid graph is achieved~\cite{queralta2020uwb}, leading to a single solution to the relative localization problem (i.e., a single localization graph realization), the null space of the transformations is defined by rototranslations of the graph realization in $\mathbb{R}^2$. 

We model UWB range measurements with Gaussian noise:
\begin{equation}
    \textbf{z}^{UWB}_{(i,j),\:(i,j)\in\mathcal{E}} = \lVert \textbf{p}_i(t)-\textbf{p}_j(t) \lVert + \mathcal{N}(0,\sigma_{UWB})
\end{equation}

Odometry egomotion estimations for each of the robots are modeled with
\begin{dmath}
    \textbf{z}^{odom}_{i,\:i\in\mathcal{V}} = \left[ \begin{array}{cc} \textbf{R}_i(t-\delta t)\hat{\textbf{R}}_i(t) & \lVert \textbf{p}_i(t)-\textbf{p}_i(t-\delta t) \lVert \\ 0 & 1 \end{array} \right] + \mathcal{N}(0,\:\sigma_{odom})
\end{dmath}
where $\delta t$ is the output frequency of the Turtlebot's odometry, $\textbf{R}_i(t)$ is the orientation matrix for agent $i$ and $\hat{\textbf{R}}_i(t)$ the relative egomotion estimation in the interval $(t-\delta t, t]$.

\begin{figure}
    \centering
    \includegraphics[width=0.35\textwidth]{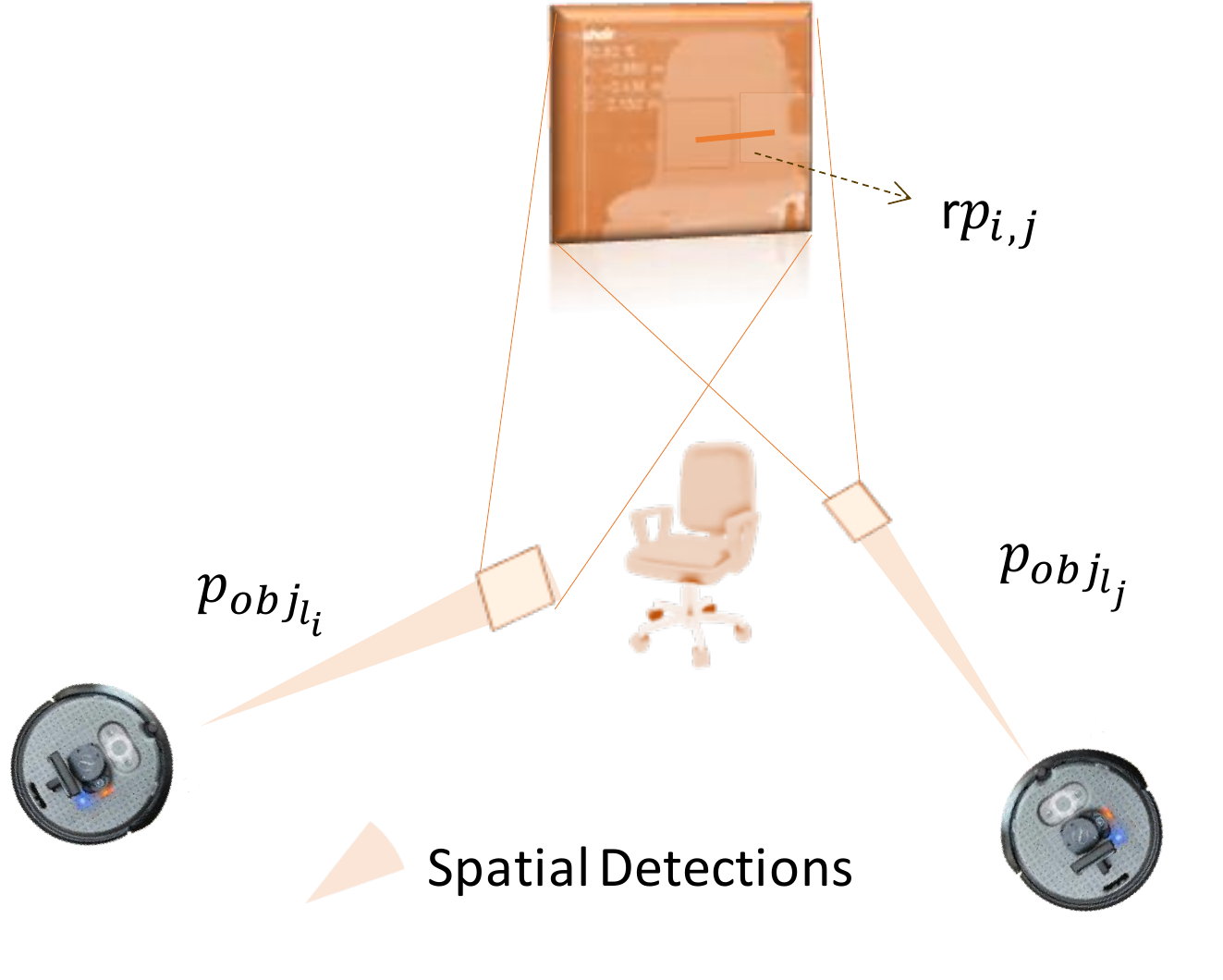}
    \caption{Cooperative spatial detection}
    \label{fig:cooperative_spatial_detection}
\end{figure}

Finally, when more than one robot detects the same objects simultaneously as shown in Fig.~\ref{fig:cooperative_spatial_detection}, we also include in the estimation the cooperative spatial detection measurements modeled as
\begin{dmath}
    \textbf{z}^{det}_{i,j} = \textbf{rp}_{i,j} + \mathcal{N}(0,\sigma_{det}) \\ \text{where} \\
    { \textbf{rp}_{i,j} = \left( \textbf{p}^{}_{obj_{l_i}}(t) + \textbf{p}_i(t) \right) - \left(\textbf{p}^{}_{obj_{l_j}}(t) + \textbf{p}_j(t) \right) }
\end{dmath}
Here, $\textbf{p}^{}_{obj_{l_i}}$ and $\textbf{p}^{}_{obj_{l_j}}$ represent the position of the simultaneously identified object relative to the reference frame of the robot in position $\textbf{p}_i$ and $\textbf{p}_j$, respectively. The vector $\textbf{rp}_{i,j} \in \mathbb{R}^2$ represents the difference between the two estimates of the position of the object made by the robot at position $\textbf{p}_i$ and $\textbf{p}_j$, respectively, with respect to the common reference frame. In other words, it represents the relative position of the object with respect to the robot at position $\textbf{p}_i(t)$ as observed from the perspective of the robot at position $\textbf{p}_j(t)$. To ensure that detections correspond to the same object, we use a distance threshold $\text{if} \:\: \lVert \textbf{rp}_{i,j} \rVert < d^{det}_{th}$ (set to \(15\,cm\) in our experiments based on the accuracy of the depth camera).

Prior to the PF process, an LSTM model was trained and employed to estimate the ranging error between each UWB pair. The details of this part can be found in section~\ref{lstm-estimation}.

\subsection{Optimization Problem Formulation}

The probability density function (pdf) of the proposal distribution indicating the importance density and denoted as \(p(S(t) | S(t-1))\) in the particle filter is modeled as a Gaussian distribution as given in Eq.~(\ref{equ:particle})

\begin{equation}\label{equ:particle}
\begin{aligned}
    S(t) &\sim \mathcal{N}(S(t-1) + \textbf{z}^{odom}(t), \mathcal{Q}(t))
\end{aligned}
\end{equation}

\noindent meaning that the global state $S_k$ at time $t$ is predicted based on a sample drawn from a Gaussian distribution with a mean vector \(S(t-1) + \textbf{z}^{odom}(t)\) and covariance matrix \(\mathcal{Q}(t)\). The state of each particle encodes a stack of 2D states of all robots, with $S(t) \in \mathbb{R}^{2N}$ and $\textbf{z}^{odom}(t) = [\textbf{z}_i^{odom}(t)]$, while $\mathcal{Q}$ is built from the odometry covariance matrix of all robots, $\mathcal{Q}_i$, $i=1,\dots, N$.

The iterative updates of the particle filter are as follows:

\noindent\textit{a) Initialization:} we initialize the states of \(M\) particle samples, \(S(t) = \{ S_k(t)\}, k \in M \) drawn from a uniform distribution given a priori knowledge of the approximate area size where robots are deployed:

\begin{equation}
   S(t=0) \sim \mathcal{U}\left( \left( x_{min}, \: x_{max} \right), \: \left(y_{min}, \: y_{max}\right) \right)
\end{equation}\label{equ:init}
with associated weights:
\begin{equation}\label{equ:weights}
     w_{k}(t=0) = \frac{1}{M}, k \in 0, \dots, M
\end{equation}

\noindent\textit{b) Prediction:} we then propagate particles forward using a motion model based on the translation from odometry data as already introduced in \eqref{equ:particle}.

\noindent\textit{c) Update (correction):} we build the observation input sample \(\textbf{z}(t)\) as a combination of inter-robot ranges $\textbf{z}^{UWB}(t)$ and spatial detections $\textbf{z}^{det}(t)$, i.e., $\textbf{z}(t) = stack [\textbf{z}^{UWB}(t), \:\textbf{z}^{det}(t)]$, with the latter being added only if available. 
Assuming we have \( 2 \lvert E \rvert \) rows of observation data, the importance weights for each particle are computed based on the likelihood of the predictions or expected measurements $\widehat{\textbf{z}}$ based on the observed data \(\textbf{z}\):
\begin{equation}\label{equ:update}
  p(\textbf{z}(t)|S_k(t)) =  \mathcal{N} \left( \widehat{\textbf{z}}_k(t) ; \textbf{z}(t), \mathcal{Q}_{obs} \right)
\end{equation}

where $Q_{obs}$ is a covariance matrix obtained from the models defining $\sigma_{UWB}$ and $\sigma_{det}$, obtained experimentally. While we do not explicitly consider a temporal component of the graph $(V,E)$, the topology might change and the method proposed in this paper is directly applicable to a variable graph structure. The predictions are calculated based on the propagated particle states by estimating virtual ranging and displacement measurements based on those states:

\begin{equation}
    \widehat{\textbf{z}}_k \longleftarrow meas\_from\_states\left(S_k \left( t \right)\right)
\end{equation}

i.e., $\widehat{\textbf{z}}_i = \{\lVert S_k[2n:2n+1] - S_k[2m:2m+1] \rVert\}_{(n,m)\in E}$ if $i < \lvert E \rvert$ and $\widehat{\textbf{z}}_i = \{ S_k[2n:2n+1] - S_k[2m:2m+1]\}_{(n,m)\in E}$ if $\lvert E \rvert \leq i < 2 \lvert E \rvert$ if a spatial detection for the same pair is available, else $nil$. The state of a robot $i$ as defined by a particle with state $S_k$ is given by $(x_i, y_i) = S_k[2i:2i+1]$.

In other words, the highest possibility in the distribution in Eq.~\eqref{equ:update} is at the location where the expected measurements $\widehat{\textbf{z}}_k$ coincide with the actual measurements \(\textbf{z}(t)\). After this, the particle weights are recalculated as follows:

\begin{equation}
    w_{k}(t)  \propto w_{k}(t-1) p(\textbf{z}(t) | S(t))
\end{equation}

\noindent\textit{d) Resampling:} in every iteration, we randomly re-initialize the particle states by sampling the inverse weight distribution. In our experiments, the resampling proportion is set to 1\%.

\noindent\textit{e) Estimation:} finally, we estimate the states based on the weighted average of the \(M\) particles. The estimated state $S(t)$ is given by:
\begin{equation}
    S(t) = \sum_{i=1}^{M} w_{k}(t) \cdot S_{k}(t)
\end{equation}

During our experiments, we compared this mean state calculation with an alternative state given simply by the particle with maximum likelihood but we observed more robust overall performance and convergence with the weighted sum.

\subsection{LSTM network for UWB ranging error estimation}\label{lstm-estimation}

\begin{figure}[t]
  \begin{subfigure}{0.15\textwidth}
    \includegraphics[width=\linewidth]{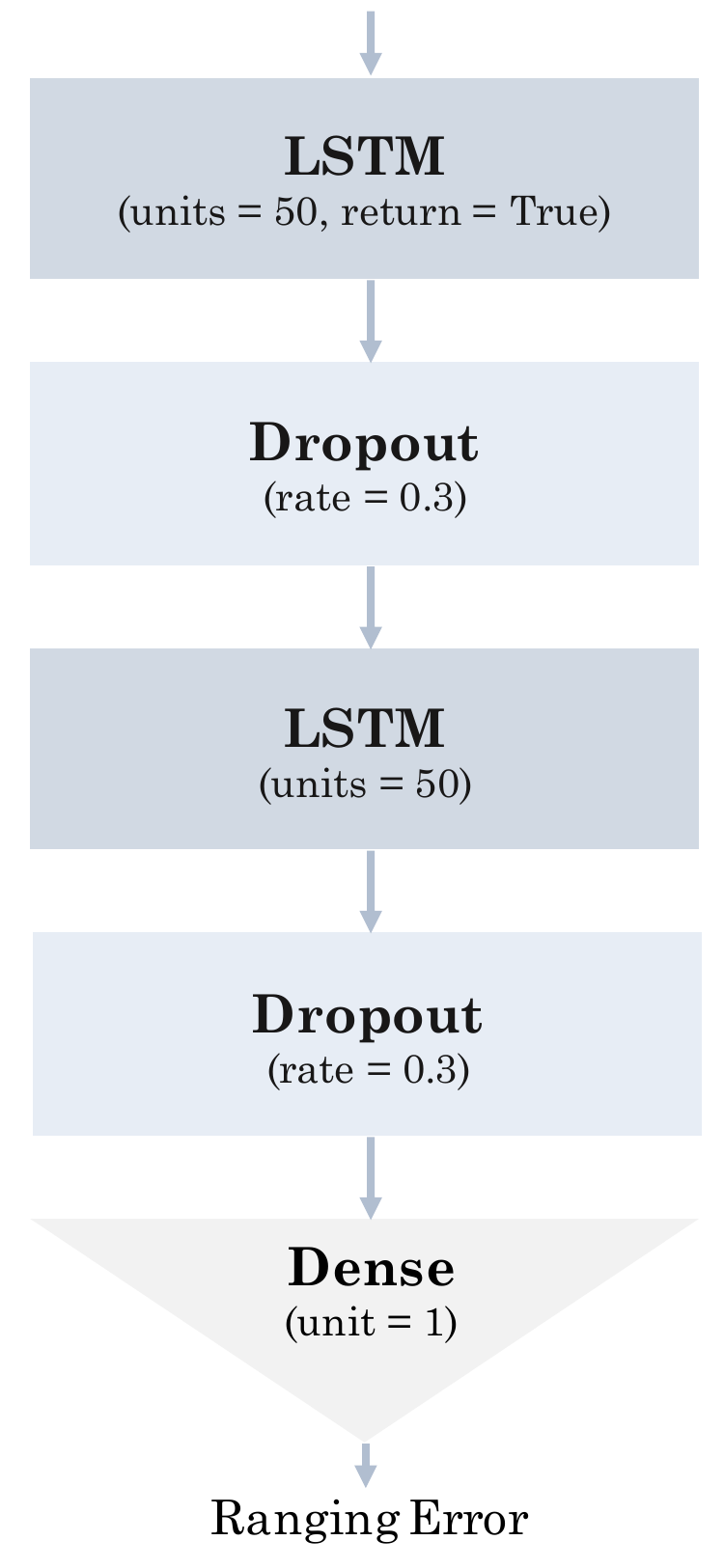}
    \caption{}
    \label{fig:sLSTM}
  \end{subfigure}%
  \hfill
  \begin{subfigure}{0.15\textwidth}
    \includegraphics[width=\linewidth]{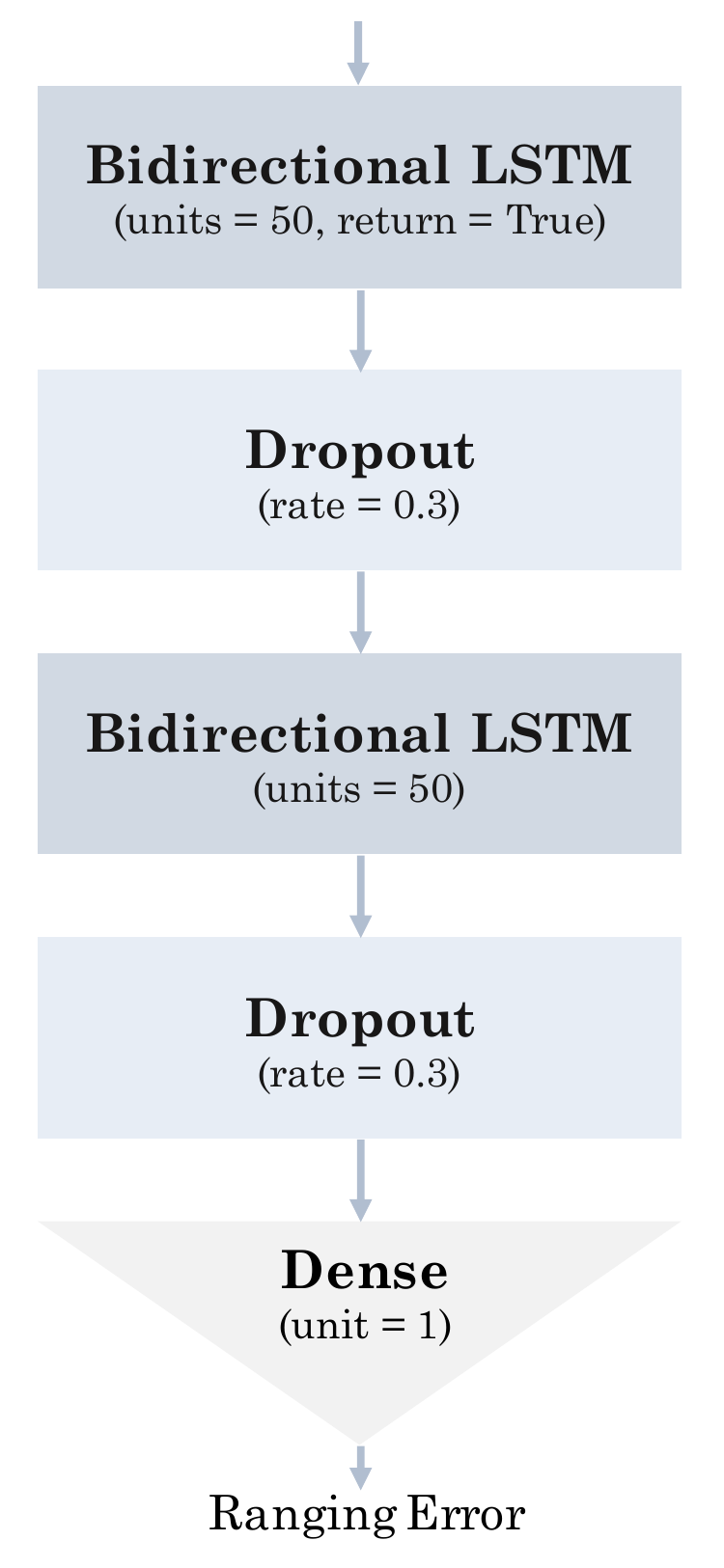}
    \caption{}
    \label{fig:biLSTM}
  \end{subfigure}%
  \hfill
  \begin{subfigure}{0.15\textwidth}
    \includegraphics[width=\linewidth]{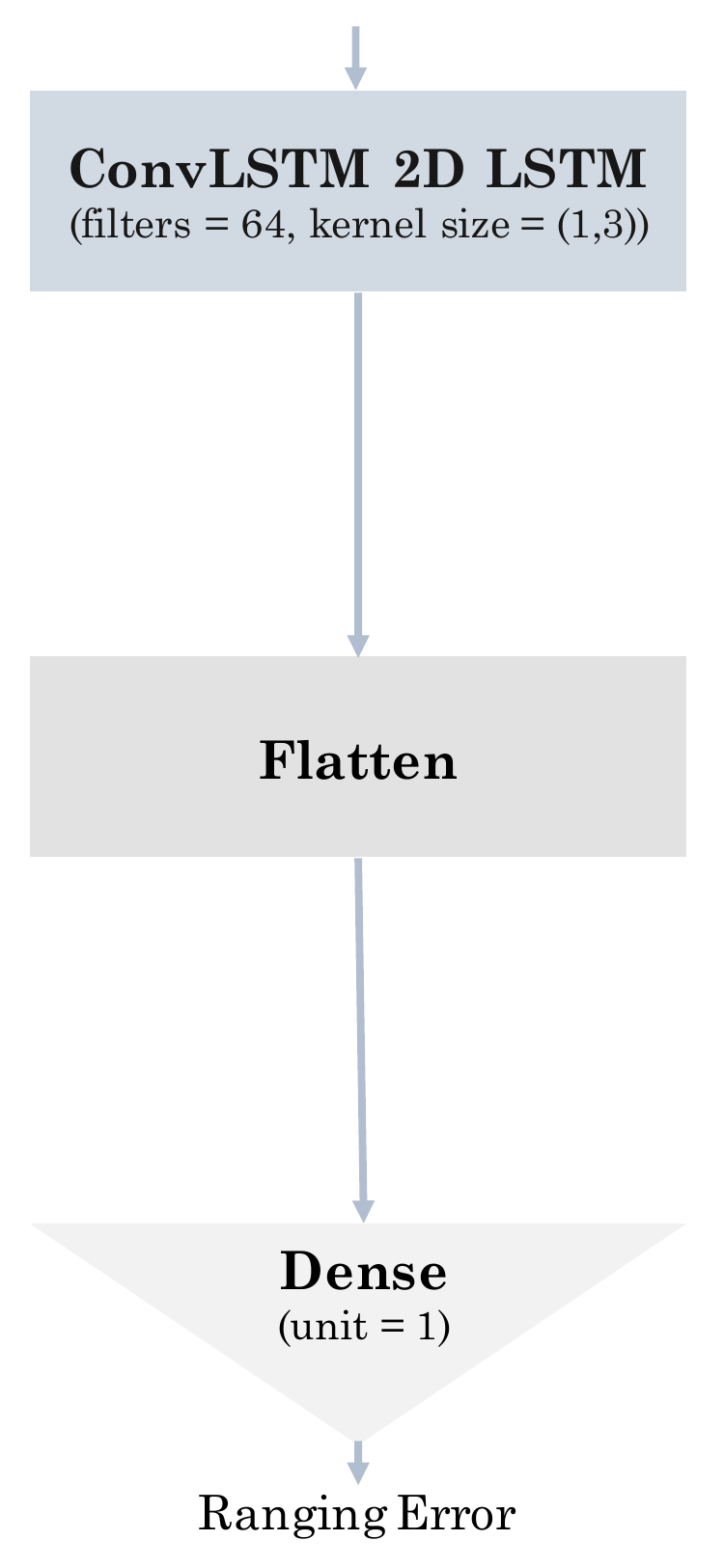}
    \caption{}
    \label{fig:conLSTM}
  \end{subfigure}
  \caption{The layer details of customized LSTM neural network variants applied in this study. Fig.~\ref{fig:sLSTM} to~\ref{fig:conLSTM} depict the implementation of stacked LSTM, bidirectional LSTM, and convLSTM, respectively.}
  \label{fig:lstms}
\end{figure}

In this study, we applied three primary LSTM variants, namely stacked LSTM, bidirectional LSTM, and convLSTM.
Each variant was designed and trained individually for each UWB pair utilized in the experiments. Fig.~\ref{fig:lstms} shows the customized layer information of these LSTM neural networks.

The inputs to these networks comprise the UWB range, denoted as $\textbf{z}^{UWB}_{(i,j),(i,j)\in\mathcal{E}}$, between the UWB pair, along with the orientations of the robots, $\theta^{odom}_i$ and $\theta^{odom}_j$, at the UWB installation ends.
To estimate the current UWB ranging error, the LSTM takes a sequence of $n\_steps$ data frames  $[\, \textbf{z}^{UWB}_{(i,j)}, \theta^{odom}_i, \theta^{odom}_j ]_{\:(i,j)\in\mathcal{E}} \,$ preceding the current time as the input. The dense layer at the end of the LSTM network generates a single UWB ranging error as the output. 

The individual LSTM networks are trained using data from a separate experiment that involved deploying the same robots within the same controlled environment and using the Optitrack ground truth system. The training data involved robots moving in circular trajectories, while the LSTM's predictions used in the state estimation were generated for different robot movement patterns as shown in Fig.~\ref{fig:harware_00}.
Therefore, we expect the networks to be able to model antenna delays and predict potential errors based on the relative orientation of the antennas. This limits the generalization of the results, but we separate in time the training data from the final experiments reported in this paper.

\subsection{Experimental setup}
\begin{figure}[t]
    \centering
    \includegraphics[width=0.45\textwidth]{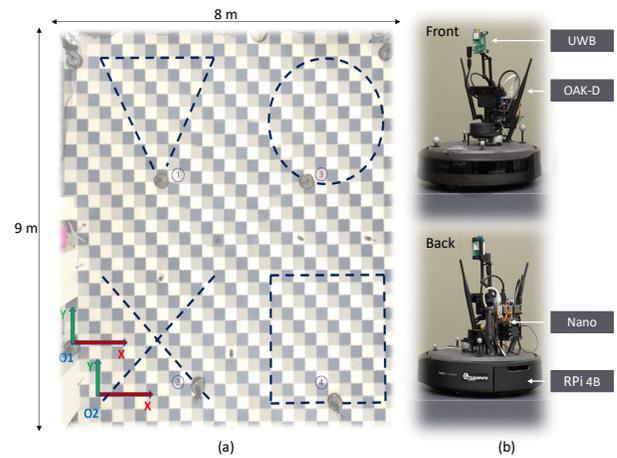}
    \caption{Experimental site and platforms. Subfigure (a) shows the motion capture (MOCAP) arena and the moving patterns of four Turtlebot4 robots. Subfigure (b) shows the customized Turtlebot4 platform mounted with a single UWB transceiver, an OAK-D stereo camera, and a Jetson Nano.}
    \label{fig:harware_00}
\end{figure}

In our experiments, we employed five Turtlebot\,4 Lite 
robots, denoted $\mathcal{TB}_{i,\:i\in\{0,1,2,3,4\}}$. These mobile robot platforms served for data collection and real-world navigation, utilizing our proposed UWB-based relative estate estimation method.  To enhance their capabilities, we customized the Turtlebot4 platform, integrating a Qorvo DWM1001 UWB transceiver and an OAK-D stereo camera, which replaced the default OAK-D Lite as shown in Fig.~\ref{fig:harware_00}\,(b).

For all our computations,  we used a Jetson Nano computer, to which the OAK-D camera is also connected, while the Turtlebot4 ROS\,2 drivers run on the existing Raspberry 4.

During the experiments, the UWB transceivers were programmed to iteratively measure the time of flight (ToF) between pairs of robots, providing the necessary ranging measurements. The OAK-D stereo cameras provide poses of the detected objects relative to the cameras as the odometry measurements are given directly by the Turthebot4 ROS\,2 drivers. 

The experimental site, as depicted in Fig.~\ref{fig:harware_00}(a) comprised an arena of approximately $8\,m\times9\,m\times5\,m$, equipped with an OptiTrack motion capture system (MOCAP) providing the ground truth to verify the estimated relative states of each robot. Within this area, we positioned a static robot $\mathcal{TB}_{4}$ and four moving robots $\mathcal{TB}_{i,\:i\in\{0,1,2,3\}}$. The static robot serves to align references with the ground truth, while the other four followed specific paths - a triangle, an $X$, a circle, and a rectangle - as shown in Fig.~\ref{fig:harware_00}(a). We assumed a common orientation frame among the robots, facilitated by a compass, enabling single-range relative state estimation. Importantly, no fixed anchor nodes were used, and the static robot's position remained unknown. 

Finally, to enable cooperative spatial detections, we utilized a pre-trained YOLOv4 network. This network effectively detected several objects, including bottles, chairs, and cups, which were placed arbitrarily both inside and outside the experimental site, serving as references.

\subsection{Multi-robot relative localization system implementation}

The system was developed using ROS 2 as the communication framework for the robots. ROS\,2 employs topics as the primary means of communication, and the Turtlebot4 is originally compatible with ROS\,2, offering a robust ecosystem. However, when operating in a multi-robot system, challenges arise in differentiating the robots for communication purposes. One approach is to use namespaces to distinguish them, but this necessitates significant modifications to the Turtlebot4's software, including the ROS\,2 navigation stack. Striking a balance between efficient communication and maintaining the robots' originality poses a considerable challenge. 

\begin{figure}[t]
    \centering
    \includegraphics[width=.45\textwidth]{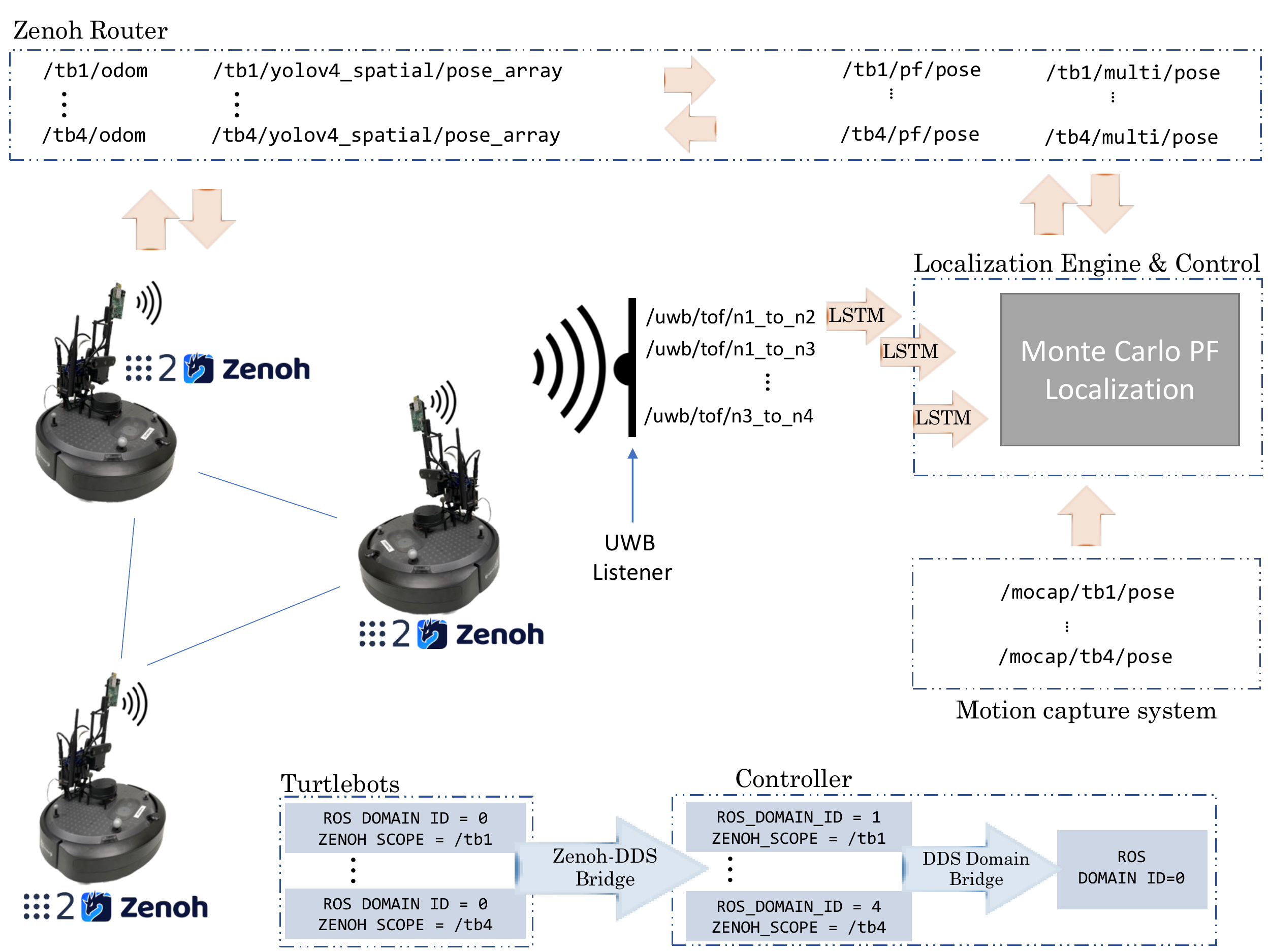}
    \caption{Implementation of the System with ROS\,2 and Zenoh}
    \label{fig:ros_zenoh_diagram}
\end{figure}

We adopted a solution that combines ROS 2 with Zenoh, while maintaining DDS as the internal middleware for each individual robot. Fig.~\ref{fig:ros_zenoh_diagram} provides a visual representation of this approach, which involves the robots, the Zenoh Router, and the localization and control engine. This engine can reside on various devices like a computer, Raspberry Pi, or Jetson Nano within the robots. In our implementation, we used a laptop for this purpose, but we demonstrate that the computational resources of any embedded computer are sufficient for the task.

We utilized Zenoh clients with different ${\scriptstyle ZENOH\_SCOPE}$ for each robot, while preserving the common DDS domain ID and topic names. This approach required no alterations to the robot settings (common DDS domain ID and topic names across robots). A Zenoh router was hosted to facilitate message discovery and transfer through the Zenoh DDS bridge\footnote{\url{https://github.com/eclipse-zenoh/zenoh-plugin-dds}}, enabling seamless communication with the localization and control engine. Ground truth data from the MOCAP system was also received by the engine.

To maintain consistency and simplicity in the communication process, each scope was mapped to a distinct domain ID at the controller end. The DDS domain bridge merged these scopes with topic namespacing\footnote{\url{https://github.com/ros2/domain\_bridge}}. Notably, the topic names remained unchanged on the receiving end (i.e., the engine), allowing for streamlined localization and control procedures. For instance, within the robots, the odometry topic was referred to as $\backslash odom$. However, in the Zenoh communication network, it became $\backslash\text{tb}\{i\}\backslash odom$ (where $\backslash\text{tb}\{i\}$ is the Zenoh scope). In the localization engine, the topic names were $\backslash odom$ with different domain IDs, or again $\backslash\text{tb}\{i\}\backslash odom$ with the common domain ID (with $\backslash\text{tb}\{i\}$ now serving as a ROS topic namespace).

Regarding the specific implementations, our PF localization is based on the filter library\footnote{\url{https://github.com/johnhw/pfilter}}, tailored to suit our program's needs.  For LSTM-based UWB ranging error estimation, we built customized LSTM models using TensorFlow Keras, performing both training and real-world inference, as illustrated in Fig.~\ref{fig:lstms}.






\subsection{Evaluation on multiple mobile computing platforms}
Given the multitude of prevalent computing platforms in contemporary multi-robot systems, presenting a performance evaluation of the proposed approach is valuable as a reference for further research in the field. 
To this end, we conducted memory consumption and CPU utilization evaluations on several platforms, including an Intel computer (i9-11900H), an NVIDIA Jetson Nano, and a Raspberry Pi 4\,B integrated with Turtlebot4. The Intel computer boasts 32\,GB of DDR4 RAM, while the Jetson Nano and Raspberry Pi 4\,B feature 4\,GB RAM. Notably, we executed the PF code without LSTM for UWB ranging error correction, as GPU support is essential for LSTM operations.

In our experiments, the Jetson Nano is needed for both spatial detections and LSTM deployment. However, the majority of the computation regarding spatial detection (both the YOLO detector and the depth data fusion) is done onboard the OAK-D camera on the robot and does not need a GPU in the companion computer.



\subsection{Navigation based on proposed relative state estimation}

To validate the relative state estimated by the proposed approach, we estimate both positioning and navigation errors in a real-world navigation task. For positioning error, the robots follow trajectories using the MOCAP data as feedback to the controller. In the autonomous navigation part, we set one robot to perform a navigation task in the shape of a rectangle using the aforementioned relative position method for controller feedback.


\section{Experimental Results}\label{sec:experiment}

We now discuss the results of different localization and navigation experiments. We analyze both single-ranging errors as well as positioning and trajectory tracking errors.

\subsection{UWB ranging error modeling}

\begin{table}[ht]
\centering
\caption{Comparison of the LSTM variants on the estimation of UWB ranging error.}
    \resizebox{\linewidth}{!}{%
        \begin{tabular}{@{}lccc@{}}
            \toprule
             & \textbf{Stacked LSTM} & \textbf{Bidirectional LSTM} & \textbf{ConvLSTM} \\
            \midrule
            \textbf{Computation Time} (\(s\)) & 0.0413 & 0.0381 & 0.0702 \\
            \textbf{MSE} (\(m^2\)) & 0.0109 & 0.008   &  0.0075 \\ \bottomrule
        \end{tabular}
    }
    \label{tb:lstm_variants}
\end{table}


Table~\ref{tb:lstm_variants} shows the Mean Square Error (MSE) and computation time of different LSTM variants, including stacked LSTM (SL), bidirectional LSTM (BL), and convLSTM (CL).


According to these results, all estimation processes can be deployed at frequencies greater than \(10Hz\) while achieving significantly lower MSE values. In particular, SL and BL can reach more than \(20Hz\) in real-time. Based on these results,  we utilized SL in the following experiments. 

\begin{figure}
    \centering
    \setlength{\figurewidth}{.48\textwidth}
    \setlength{\figureheight}{.30\textwidth}
    \scriptsize{\input{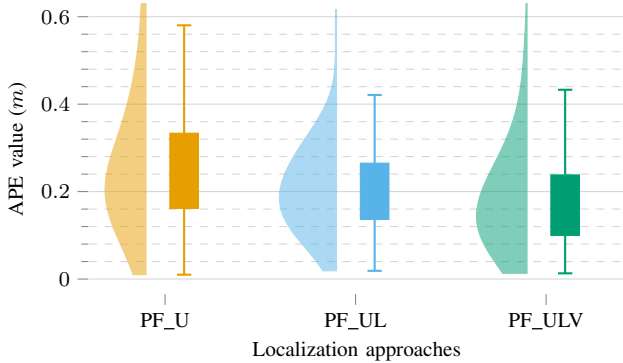}}
    \caption{Absolute Positioning Error (APE) for the two-robot, single-range positioning experiment.}
    \label{fig:single_ape}
\end{figure}

\subsection{Relative State Estimation based on a Single UWB range}

We evaluated the proposed relative state estimation approach for a single UWB range considering the Absolute Positioning Error (APE) and the trajectory of the ground truth. In the results, we denote the particle filter with only UWB ranges, stacked LSTM corrected UWB ranges, and both corrected UWB ranges and dynamic cooperative spatial detections by PF\_U, PF\_UL, and PF\_ULV, respectively.

\begin{figure}
    \centering
    \setlength{\figurewidth}{0.45\textwidth}
    \setlength{\figureheight}{0.45\textwidth}
    \footnotesize{\input{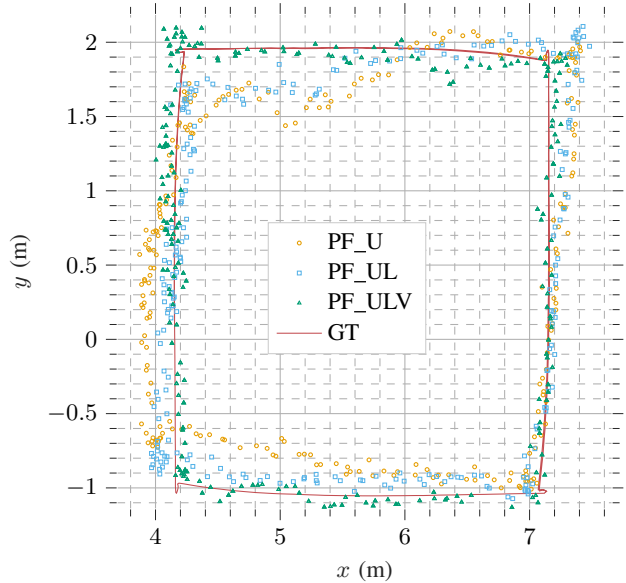}}
    \caption{Trajectories of the state estimation with a single UWB range.}
    \label{fig:single_traj}
\end{figure}

For the specific experiment involving a single UWB range measurement between two robots, multilateration methods cannot calculate the relative position between the robots.  
Therefore, in this part, we only analyze the APE values and the trajectories of PF\_U, PF\_UL, and PF\_ULV. 
According to Fig.~\ref{fig:single_ape}, we can improve the proposed PF by mitigating the UWB ranging error with LSTM. Moreover, by integrating cooperative spatial detection, we can further enhance its performance.
For a tangible comprehension, fig.~\ref{fig:single_traj} shows the trajectories of these proposed methods.

\begin{figure}[b]
    \centering
    \setlength{\figurewidth}{0.48\textwidth}
    \setlength{\figureheight}{0.48\textwidth}
    \scriptsize{\input{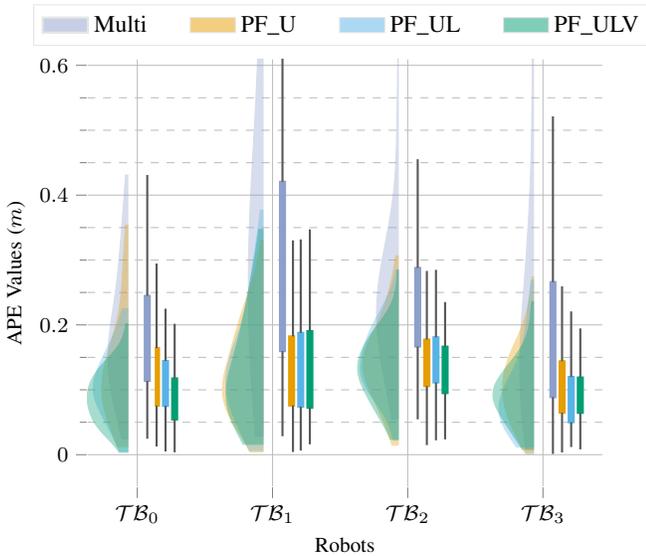}}
    \caption{Absolute Positioning Error (APE) values of the four relative localization methods with four robots moving in four distinct patterns.}
    \label{fig:multi_ape}
\end{figure}

\subsection{Relative state estimation for multiple UWB ranges}

\begin{figure}
    \centering
    \setlength{\figurewidth}{.48\textwidth}
    \setlength{\figureheight}{.48\textwidth}
    \scriptsize{\input{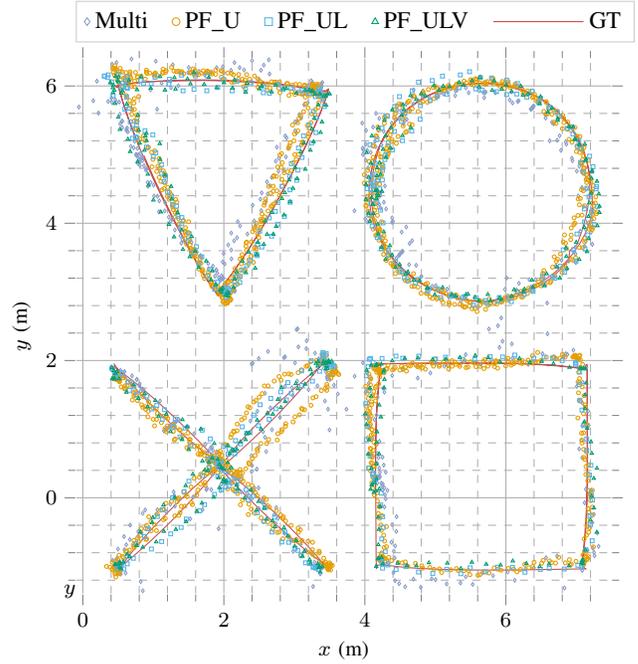}}
    \caption{Trajectories of robots moving in distinct patterns with different relative state estimation approaches.}
    \label{fig:multi_tracj}
\end{figure}

Fig.~\ref{fig:multi_ape} shows the APE values of the four approaches with \textit{Multi} denoting multilateration applied to robots moving in four distinct patterns. 
It is evident from the figure that our proposed PF approach significantly outperforms multilateration. Moreover, incorporating the LSTM network and integrating cooperative spatial detection can improve the performance of the proposed PF in most cases. To give a concrete idea, Fig.~\ref{fig:multi_tracj} depicts the trajectories of the relative localization approaches.

\subsection{Resource utilization}

Table~\ref{tb:resources_consumption} shows the CPU usage of the proposed methods on Intel PC, Nano, and Raspberry Pi. The CPU utilization for the programs running on Nano and Raspberry Pi is around \(20\%\), which is relatively low. Additionally, the memory consumption of the proposed approach is always below \(500M\) across all the platforms. It is worth noting that the program running on the computing platforms is to calculate the relative states of all robots instead of just each individual state.


            

\begin{table}[h]
\centering
\caption{CPU utilization of the different methods.}
    \resizebox{\linewidth}{!}{%
    \begin{tabular}{@{}lccc@{}}
        \toprule
         & \textbf{Intel PC (i9-11900H)} & \textbf{Jetson Nano 4GB} & \textbf{Raspberry Pi 4B 4GB} \\
        & \multicolumn{3}{c}{\textit{( CPU (\%), RAM (MB))}} \\
        \midrule
        \textbf{Multilateration} & (2.9, 390) & (21.3, 99) & (22.4, 66) \\
        \textbf{Our Approach} & (2.8, 69) & (23.2, 442)   &  (23.2, 83) \\ \bottomrule
    \end{tabular}
    }
    \label{tb:resources_consumption}
\end{table}

\subsection{Navigation performance}
In real-world navigation performance assessment, we compute the median and standard deviation of the APE and Absolute Trajectory Error (ATE). The APE quantifies positioning error, while the ATE assesses the robot's ability to track a prefined path using the proposed approach. Specifically, the APE for positioning error is \( 0.1094\,m/0.10125\,m\), and the ATE value is approximately \(0.0872\,m/0.1011\,m\).

\section{Conclusion}\label{sec:conclusion}

We have presented a particle filter-based relative multi-robot localization approach, that fuses inter-robot UWB ranges, robot odometry, and cooperative spatial detections. Contrary to conventional approaches such as multilateration, our method can estimate the relative position even from a single UWB sensor. We also experiment with LSTM networks trained for each UWB pair to predict the ranging error, which has proven to be accurate and reliable in real-time. The accuracy and real-time features show the potential of applying the LSTM network to every UWB-ranging measurement before feeding them to the proposed particle filter. Additionally, our approach dynamically integrates cooperative spatial detections from the mounted stereo cameras when available. Specifically, as long as two or more robots detect identical external objects, the proposed particle filter dynamically takes the extracted spatial information between the two robots as inputs. According to the experimental results, our approach clearly outperforms multilateration for relative state estimation.
Furthermore, our findings show that the LSTM error estimation and spatial information extracted by our cooperative method can improve the performance with respect to using only UWB data. Finally, we evaluated our approach and program with ROS\,2 and Zenoh on several popular mobile computing platforms which indicates our approach has low CPU utility and memory consumption.



\section*{Acknowledgment}

This research work is supported 
by the Academy of Finland's 
AeroPolis project (Grant No. 348480)
, and by the R3Swarms project funded by the Secure Systems Research Center (SSRC), Technology Innovation Institute (TII).

\bibliographystyle{IEEEtran}
\bibliography{bibliography}
\end{document}